\documentclass[a4paper,fleqn]{cas-sc}

\usepackage[numbers]{natbib}
\usepackage{graphicx}
\usepackage{blindtext}
\usepackage{float}
\usepackage{multicol}
\usepackage{subfig}
\usepackage{caption}
\usepackage{multirow}
\usepackage{booktabs}

\begin{document}
\let\WriteBookmarks\relax
\def\floatpagepagefraction{1}
\def\textpagefraction{.001}

\shorttitle{Selecting Update Blocks of Convolutional Neural Networks using Genetic Algorithm in Transfer Learning}

\shortauthors{Mehedi Hasan et~al.}


\title [mode = title]{Speeding Up EfficientNet: Selecting Update Blocks of Convolutional Neural Networks using Genetic Algorithm in Transfer Learning}

%

\author[1]{Md. Mehedi Hasan}
\address[1]{Dept. of Computer Science and Engineering, East West University, Dhaka, Bangladesh}
\cortext[1]{Corresponding author}
\ead{mehedi.mailing@gmail.com}
\author[2]{Muhammad Ibrahim*} [orcid=0000-0003-3284-8535]
\address[2]{Dept. of Computer Science and Engineering, Univeristy of Dhaka, Dhaka, Bangladesh}
\ead{ibrahim313@du.ac.bd}
\author[1]{Md. Sawkat Ali}
\ead{alim@ewubd.edu}





\begin{abstract}
The performance of convolutional neural networks (CNN) depends heavily on their architectures. Transfer learning performance of a CNN relies quite strongly on  selection of its trainable layers. Selecting the most effective update layers for a certain target dataset often requires expert knowledge on CNN architecture which many practitioners do not posses. General users prefer to use an available architecture (e.g. GoogleNet, ResNet, EfficientNet etc.) that is developed by domain experts. With the ever-growing number of layers, it is increasingly becoming quite difficult and cumbersome to handpick the update layers. Therefore, in this paper we explore the application of genetic algorithm to mitigate this problem. The convolutional layers of popular pre-trained networks are often grouped into modules that constitute their building blocks. We devise a genetic algorithm to select blocks of layers for updating the parameters. By experimenting with EfficientNetB0 pre-trained on ImageNet and using Food-101, CIFAR-100 and MangoLeafBD as target datasets, we show that our algorithm yields similar or better results than the baseline in terms of accuracy, and requires lower training and evaluation time due to learning less number of parameters. We also devise a metric called block importance to measure each block's efficacy as update block and analyze the importance of the blocks selected by our algorithm.
\end{abstract}




\maketitle

\section{Introduction}
High-performance image recognition models are often developed using Convolutional Neural Networks (CNN). As the prevalent approach of deep learning 
in image classification, CNNs have shown exceptional supremacy over many approaches in various real-world machine-learning applications \cite{Krizhevsky2012}. The performance of a CNN depend heavily on its architecture \cite{ShimonyanZisserman2015}, and hence, all of the state-of-the-art CNNs, such as GoogleNet \cite{Szegedy2015}, ResNet \cite{HeZhang2016}, DenseNet \cite{HuangLiu2017} etc., are handcrafted by experts who have rich domain knowledge. As it is not always feasible for general practitioners of CNN to acquire such expertise, these users often opt to use a pre-designed architecture that suits their need. CNNs are usually designed at a fixed computational resource budget, and then scaled up at a later time for better accuracy if more computational resources become available.

The training process of CNNs requires large sized datasets because these models need to learn a huge number of parameters. Since the parameter space is colossal, sufficient amount of training data are warranted to learn complex patterns. This requirement of having large datasets, however, can be relaxed using the transfer learning setting \cite{PanYang2010} where practitioners reuse existing pre-trained networks, thereby reducing the training time significantly. In transfer learning, during training phase, the parameters of some layers of the pre-trained network are kept fixed while updating the rest with the (target) dataset at hand. 
Generally, early layers (near the input) of a CNN detect low-dimensional information like the color and edges of an image, and the later layers (near the output) extract high-dimensional features that help to identify the ground truth labels \cite{ZeilerFergus2014}. Therefore,  for transfer learning, usually the early layers of the pre-trained model are frozen and parameters of the later layers are updated.

Some recent empirical studies, however, report good results by applying the opposite practice, i.e., keeping fixed the parameters of the later layers instead of that of earlier ones. To mention a few such works: Zunair et al \cite{ZunairMuhammedMomen2018} use a VGG16 network \cite{ShimonyanZisserman2015} pre-trained with ImageNet for the prediction task of Bangla characters and report that the best accuracy is found when the first input layer and early fully-connected layers are selected as update layers. Gafoorian et al \cite{Ghafoorian2017} apply transfer learning on MRI images and report that the best performance is found when only six input layers are updated. Therefore, it is intriguing to investigate into the appropriate  layers to be updated for transfer learning.


Nowadays, common CNN architectures have a lot of layers. For example, VGG16, InceptionV3, and GoogleNet have 16, 94, and 22 layers respectively. EfficientNetB0 and EfifcientNetB7 \cite{TanLe2019} have 237 and 813 layers respectively. When the general users want to use a pre-trained model, such large number of layers  makes it quite hard for them to manually select the appropriate layers to be updated. Manually selecting the layers to be updated involves a trial-and-error approach which is time consuming and tiresome.

To mitigate this difficulty, there are studies that apply the so-called metaheuristic optimization algorithms (such as genetic algorithm) to automatically select the effective layers to be updated (we discuss these works detail in Section~\ref{sec:related work}). However, we have not found any work that investigates into selecting the appropriate blocks -- not individual layers -- to be updated in a transfer learning setting using genetic algorithms. Our investigation is dedicated to this endeavor.

In this paper, 
we develop a genetic algorithm-based method to automatically select blocks of layers -- instead of individual layers -- that significantly minimizes the training time of CNNs while maintaining similar accuracy. We also adapt a recently proposed metric named OTDD \cite{DavidNicolo2020} to calculate importance of blocks of layers. Using it we calculate the contribution of each block's in identifying the features. In all these investigations we use three target datasets, namely, Food-101, CIFAR-100, and MangoLeafBD. As for the CNN, we use EfficientNet \cite{TanLe2019} models pre-trained on ImageNet~\cite{Deng2009}.



The rest of the paper is organized as follows. Section~\ref{sec:background study} presents the background knowledge required to understand the paper.   Section~\ref{sec:proposed} presents the framework and methodology of the investigation. Section~\ref{sec:results} demonstrates the experimental results. Section~\ref{sec:related work} discusses the relevant existing works. Finally, Section~\ref{sec:conclusion} concludes the paper.

\section{Background Study}
\label{sec:background study}

As mentioned earlier, in this investigation we use EfficientNet \cite{TanLe2019} models pre-trained on ImageNet~\cite{Deng2009}, though our developed framework is equally applicable to other types of CNNs as well. In this section we briefly discuss the architecture of EfficientNet and PathNet. We also discuss a metric called Optimal Transport Dataset Distance (OTDD). 


\subsection{EfficientNet}
CNNs are often scaled up to achieve better performance. For example, ResNet \cite{HeZhang2016} can be scaled up from ResNet-18 to ResNet-200 by incorporating more layers. GPipe's \cite{HuangCheng2018} ImageNet top-1 accuracy is improved to \text{84.3\%} by scaling up the baseline model by 4 times. Scaling CNNs up by their depth \cite{HeZhang2016} or width \cite{Zagoruyko2016} are the most commonly performed, but scaling up models by image resolution \cite{HuangCheng2018} is also a popular method. Only one of the three dimensions -- depth, width, and image size -- is usually scaled. Although two or three dimensions can be arbitrarily scaled, it involves tedious manual tuning and still often fails to provide better accuracy and efficiency. To resolve this, EfficientNet proposes a simple yet effective compound scaling method that uses a constant scaling ratio to scale all three dimensions of network in a controlled and balanced way. The intuition is: as the input image size increases, it makes sense that the model will need more layers and more channels to extract even finer details from larger images. In fact, previous studies \cite{RaghuPoole2017}, \cite{LuPu2018}, \cite{Zagoruyko2016} have shown that there is a certain correlation between network width and depth. The compound scaling method uniformly scales network width, depth, and resolution with a set of constant scaling coefficients. In particular, if \(2^N\) times more computational resources become available, then this method simply scales up the network depth by \(\alpha^N\), width by \(\beta^N\), and image size by \(\gamma^N\), where \(\alpha, \beta, \gamma\) are constant coefficients chosen through a small grid search on the baseline network.

\subsection{PathNet and StepwisePathNet}

When performing transfer learning to the target datasets, each layer of a pre-trained CNN detects features that are common among the source and target datasets \cite{ZeilerFergus2014}. Therefore, it is imperative to efficiently select layers that are effective feature detectors for the target datasets and then update their parameters. Additionally, as the network architectures have become more complex due to the increased availability of computational resources, an efficient way of selecting effective layers without manual labor is required. The StepwisePathNet method \cite{ImaiKawai2020} tries to address this specific need. This algorithm expands DeepMind's PathNet \cite{FernandoBanarse2017} to select the update layers in a straight-chain network. StepwisePathNet algorithm labels each layer as either fixed or updated, and the selection is optimized by a genetic algorithm that employs a tournament selection mechanism. 

\subsection{Improvement on StepwisePathNet}

Citing the limitations of StepwisePathNet, Nagae et al~\cite{Nagae2022} improves the algorithm by applying another genetic algorithm. The authors work with InceptionV3 architecture \cite{szegedy2016rethinking} and apply a genetic algorithm to automatically select the effective layers of the network to be updated during learning for the target dataset.

\subsection{Optimal Transport Dataset Distance (OTDD)}
Evaluation of the distance between two labeled datasets has been explored in studies utilizing the optimal transport distance \cite{DavidNicolo2020}, which provides a way to quantify the difference between two datasets and correlate dataset distance with transfer learning efficacy. Optimal transport deals with the issue of transferring material from one place to another at minimum cost, and can also be used to mathematically compare two different probability distributions \cite{Villani2008}. 

Optimal Transport Dataset Distance (OTDD) \cite{DavidNicolo2020} is a metric to compute the distance between two different labeled datasets. Using this metric, in \cite{Nagae2022}, the importance of a layer is calculated by estimating its effectiveness as an update layer for transfer learning and its potential for detecting common traits in both source and target datasets. A subset of feature maps generated at layer $l$ for both source and target datasets are taken and denoted respectively as $A_l^{source}$ and $A_l^{target}$. Then, the layer importance $LI$ is expressed as:
\begin{equation}
\label{eq:22}
LI(l) = \frac{OTDD(A_l^{source}, A_l^{target})}{OTDD(A_l^{source}, A_l^{source^\prime}) + \epsilon}
\end{equation}

Here, OTDD is calculated using Equation~16 of \cite{Nagae2022} and $\epsilon$ is a small number. The denominator in Eq.~\ref{eq:22} denotes the optimal transport distance between two different subsets of the source dataset, where the difference is created due to the difference in sampling. This ratio basically captures the difference between the features maps generated for the source and target datasets at a paricular layer. Datasets with similar feature sets should result in similar feature maps at an effective layer, yielding a lower $LI$. A lower value of $LI$ indicates higher adaptability of the model for the target dataset.

\section{Proposed Framework and Methodology}
\label{sec:proposed}

It is well-known to the research community that the performance of CNNs is highly dependent on their architecture, and hence all high performing CNNs like GoogleNet, ResNet, DenseNet, EfficientNet, InceptionV3 etc. have been manually designed by experts who possess profound  knowledge on CNNs. Unfortunately, such deeper understanding of CNNs and expertise in machine learning cannot be expected from all general users. Hence, general users often opt to use a pre-designed architecture that suits their need, thereby rising the notion of transfer learning. In this setting, the user of a CNN does not need to train the model on a large dataset, and instead  takes advantage of a pre-trained model which has already been trained on a large source dataset. The intuition is as follows. Some layers of CNNs extract low level information such as edge, color, shape etc. from input images, while other layers detect high dimensional features like ground truth labels of the instances. Since in image classification task, the edges and low-level shapes of images are needed to be extracted irrespective of the domain at hand, there is no benefit to re-extract these features, and so the parameters that contribute to extracting these low level features are no longer needed to be re-learnt across different domains/datasets. Therefore, in a transfer learning setting, it is imperative to decide which layers should be kept fixed (i.e., no new learning of parameters are needed) and which layers should be learnt/updated anew for the target dataset. The challenge, however, is, as the depth and complexity of CNNs are rapidly growing with the increasing availability of computational resources, it is becoming infeasible for general practitioners to handpick effective layers for update. 

Many popular CNN architectures such as Resnet, MnasNet, GoogleNet, EfficientNet etc. are constituted of groups of convolutional layers which are called blocks. Each group of layer or block helps the model identify a low or high dimensional characterizing feature. From this intuition, we pose the research question: ``\emph{Can we automatically select appropriate blocks for update using a genetic algorithm that would yield at least similar accuracy to the baseline model}?'' The effect will be  lesser blocks in the network (i.e., reduced number of parameters) to be learnt for transfer learning, thereby reducing the execution time.

\subsection{Automatic Block Selection for Update by Genetic Algorithm}
In our proposed scheme, a genetic algorithm is devised to select the blocks to be updated so as to reduce the training time, and, at the same time, to yield good   accuracy in prediction for the target datasets. A genetic algorithm is a metaheuristic search algorithm that selects a good-enough solution from the vast search space of potential solutions. It trades off between exploration and exploitation, which means, optimizing a potential solution while escaping the local minima. This algorithm, broadly, works as follows. It begins its journey in the solution space with some potential solutions whose set is called ``population''. It then selects two ``parent'' solutions from the solution pool based on some fitness function, and then applies two operations, namely crossover and mutation, to generate ``children'' solutions. This process is repeated until a good-enough solution is found. Use of the fitness function ensures exploitation of the search space, and use of randomization allow exploration of the search space.

In our method, we maintain a binary array to denote the blocks of the network where each block is denoted by 0 or 1. A 1 means the block is selected update. Thus, for each solution or genotype $g$, if the $i$th gene $g_i$ is equal to $1$, then the corresponding $i$th block of the network, i.e., the feature extracted model, is selected for update, which means, all of the layers of this block are selected as update layers. If the block is not selected, then its layers are frozen, i.e., the parameters of its layers are not updated. Then, the obtained model is trained for a single epoch on the target dataset, and the test accuracy of this model on the evaluation data is regarded as the fitness of genotype $g$. This process is fine-tuned for 100 rounds (known as epochs) to obtain the final model. 

\subsection{Block Accuracy and Block Importance}
In order to understand the impact of each block on accuracy of the target dataset, we evaluate the obtained test accuracy of the fine-tuned model on target datasets when only a particular block is selected to be updated. For each block, only that particular block in the feature extracted model is selected to be updated and the rest are frozen. The model is then fine-tuned on the target dataset for up to 20 epochs, and the evaluation accuracy of the final model is recorded as the block accuracy denoted by $BA$.

Activation feature map datasets for all the source and target datasets are generated for each block of the pre-trained model, which are then used to calculate the Optimal Transport Dataset Distance (OTDD) between the pairs of source and target datasets.
The OTDD is calculated according to the method of \cite{DavidNicolo2020} using their implementation\footnote{\url{https://github.com/microsoft/otdd}}.


Following the definition of layer importance (defined in Equation 13 of \cite{Nagae2022}), we define block importance, $BI$ for $b$th block as follows:
\begin{equation}
\label{eq:23}
BI(b) = \frac{OTDD(A_b^{source}, A_b^{target})}{OTDD(A_b^{source}, A_b^{source^\prime}) + \epsilon}
\end{equation}

This ratio adeptly captures the difference between the feature maps generated for the source and target datasets at a paricular block. Datasets with similar feature sets should result in similar feature maps at an effective block, yielding a lower $BI$. Lower value of $BI$ indicates higher adaptability of the model for target dataset.


\subsection{Model}
We apply our proposed scheme on a popular pre-trained network for transfer learning called EfficientNetB0 which has around 237 layers grouped in 8 blocks. For larger models of the EfficientNet family, similar results are expected since they are just scaled up versions of the base model and the basic building blocks remain the same. Moreover, since our method is generic, we expect similar result on other pre-trained networks.

As mentioned earlier, EfficientNetB0 is pre-trained on ImageNet dataset. As target datasets, we use three datasets, namely CIFAR-100, Food-101, and MangoLeafBD (details of these datasets are discussed in the next section). 
Instead of selecting individual layers (as done in \cite{Nagae2022}), our genetic algorithm select blocks of layers for update blocks from the feature-extracted model. Fully connected layers of the model are always selected as update layers.

We set the population size to 7. Elite plus roulette method is used as the selection method. The mutation probability is set to 1\%.
The training phase is performed  for 100 epochs with the Adam optimizer (0.0001 learning rate). The best model obtained from the genetic algorithm is then trained on the target dataset. The training accuracy, validation accuracy, training time and the number of parameters are recorded as evaluation metrics.


\section{Experimental Results}
\label{sec:results}
In this section we describe the experimental results and findings. We measure the performance against five parameters: update layer/block selection time, training time, evaluation time, accuracy, and number of parameters to be learnt.

\subsection{Datasets and Experimental Settings}
We use three target datasets: (1) CIFAR-100 \cite{KrizhevskyHinton2009}, (2) Food-101 \cite{Bossard2014}, (3) MangoLeafBD \cite{mangodata}. CIFAR-100 is an object recognition dataset with 100 classes each having 500 images for training and 100 images for testing. Food-101 is a food recognition dataset with 101 classes each of which has 750 training and 250 testing images. MangoLeafBD is a recenly released dataset containing 4000 images of mango leaves with 8 classes of diseases. All input images are shuffled, resized to $224 \times 224$ pixels, batched and pre-fetched for optimal data loading and training performance. All experiments are performed in a single NVIDIA GeForce RTX 2060 with a batch size of 32. 

\subsection{Baseline: Automatic Layer Selection by Genetic Algorithm}
As the baseline method, we compare our method with the work of Nagae et al. \cite{Nagae2022}. Here a genetic algorithm is used to select the appropriate update layers for the target datasets. 

\subsection{Performance Comparison}
Table~\ref{tab:alldata_performance} presents the experimental results  of Food-101, CIFAR-100, and MangoLeafBD target datasets.

Here LayerSelect is the layer selection algorithm by \cite{Nagae2022}, and BlockSelect is our proposed algorithm.    In the 1st column, runtime means the time taken by the layer/block selection algorithm, training time is the networks' training time after selecting layers/blocks, Evaluation time means the testing time, accuracy is the percentage of correct prediction on test set, and finally \# parameters is the number of parameters to be learnt for learning the target datasets.

\begin{table}[width=\linewidth,cols=4,pos=h]
\caption{Performance on three datasets. \emph{LayerSelect} stands for the layer selection algorithm by \cite{Nagae2022} and \emph{BlockSelect} means our proposed block selection scheme.}
\label{tab:alldata_performance}
\begin{tabular*}{\tblwidth}{ c | c | c | c | c | c | c }
\toprule
& \multicolumn{2}{c|}{Food-101}  & \multicolumn{2}{c|}{CIFAR-100} & \multicolumn{2}{c}{MangoLeafBD} \\
\toprule
 & LayerSelect & BlockSelect & LayerSelect & BlockSelect & LayerSelect & BlockSelect\\
\midrule
Runtime       & 3 hours & 43 mins & 1h 34m & 39 mins & 12m 19s & 2m 42s\\
Training time     & 31 mins     & 21 mins     & 28 mins     & 19 mins     & 19 mins     & 18 mins\\
Evalation time & 42 ms & 32 ms & 31 ms & 31 ms & 58 ms & 99 ms\\
Accuracy     & 0.77    &  0.79     & 0.81    &  0.82    & 1.0    &  0.997 \\
\# parameters & 5,79,813 & 31,13,413 &  2,56,500 & 8,30,758 & 3,61,896 & 4,24,784\\
\bottomrule
\end{tabular*}
\end{table}

Our proposed block selection algorithm works much faster than the baseline, i.e., the layer selection algorithm \cite{Nagae2022}. Still we achieve slighly better results in two datasets and slightly worse result in the other dataset. Overall, the block selection algorithm maintains the similar level of accuracy while reducing the training and evaluation time.

From the experimental results it is evident that the block selection algorithm is faster than the layer selection method and yields a model that has similar or better accuracy and inference time.

\subsection{Block Importance and Block Accuracy}
\begin{figure}
	\centering
		\includegraphics[scale=2.0]{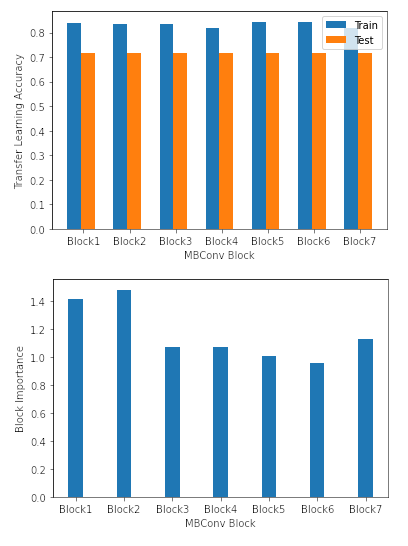}
	\caption{Block importance (BI) and block accuracy (BA) of various blocks for Food-101 dataset.}
	\label{FoodBI}
\end{figure}

\begin{figure}
	\centering
		\includegraphics[scale=0.5]{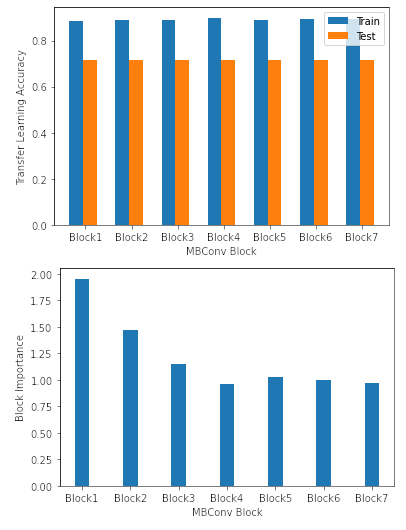}
	\caption{Block importance (BI) and block accuracy (BA) of various blocks for CIFAR-100 dataset.}
	\label{CIFARBI}
\end{figure}

\begin{figure}
	\centering
		\includegraphics[scale=0.7]{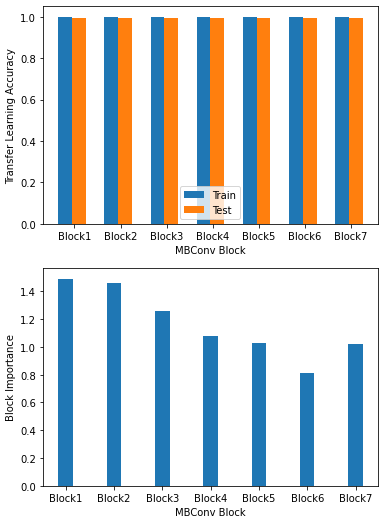}
	\caption{Block importance (BI) and block accuracy (BA) of various blocks for MangoLeafBD dataset.}
	\label{MANGOBI}
\end{figure}

Table~\ref{tab:BI_BA_alldata} presents the block importance (BI) and block accuracy (BA) metrics found for the target datasets, and 
Figs.~\ref{FoodBI}, \ref{CIFARBI}, and \ref{MANGOBI} illustrate the results graphically. Block importance is calculated using Equation~\ref{eq:23}. Block accuracy is the obtained from the training and testing accuracy  when transfer learning is performed with only considering the update layers of that block.

\begin{table}[width=\linewidth,cols=4,pos=h]
\caption{Block importance (BI) and block accuracy (BA) of various blocks for all three datasets.}
\label{tab:BI_BA_alldata}
\begin{tabular*}{\tblwidth}{ c | c | c | c | c | c | c | c | c | c }
\toprule
& \multicolumn{3}{c|}{Food-101}  & \multicolumn{3}{c|}{CIFAR-100} & \multicolumn{3}{c}{MangoLeafBD} \\
\toprule
 Block & BI & Train BA & Test  BA & BI & Train  BA & Test  BA & BI & Train  BA & Test  BA \\
\midrule
1 & 1.420 & 0.84 & 0.72 & 1.952 & 0.88 & 0.72 & 1.491 & 1 & 0.993\\
2 & 1.484 & 0.83 & 0.72 & 1.471 & 0.89 & 0.72 & 1.459 & 1 & 0.995\\
3 & 1.073 & 0.83 & 0.72 & 1.152 & 0.89 & 0.72 & 1.259 & 1 & 0.995\\
4 & 1.072 & 0.82 & 0.72 & 0.964 & 0.90 & 0.71 & 1.078 & 1 & 0.995 \\
5 & 1.011 & 0.84 & 0.72 & 1.021 & 0.89 & 0.72 & 0.029 & 1 & 0.995 \\
6 & 0.959 & 0.84 & 0.72 & 1.000 & 0.90 & 0.71 & 0.815 & 1 & 0.995 \\
7 & 1.132 & 0.82 & 0.72 & 0.965 & 0.89 & 0.72 & 0.022 & 1 & 0.995 \\
\bottomrule
\end{tabular*}
\end{table}

Conventionally for transfer learning, it is believed that updating layers close to the output side of the network is more effective. From our experimental data we cannot decisively claim this conjecture to be true, but it is evident that updating output side blocks work pretty well. We also see that updating blocks with lower BI results in high training and testing accuracy. Though we cannot conclusively claim that updating the blocks with higher BI does not work, the trends shown by the experimental data evidently suggests that updating blocks with lower BI is effective for transfer learning.

\section{Related Work}
\label{sec:related work}

Deep learning and transfer learning are well-known for their effectiveness in image classification task \cite{strawberry}. It is well-known that the structure and performance of a CNN rely heavily on its hyper-parameters. Hyper-parameters are often manually chosen by experts to obtain a model with expected performance. However, different datasets may require different model structure, and hence, choosing them by trial and error can be tedious. \citet{AszemiDomic2019} discuss the usefulness of different search and optimization methods to find a suitable set of hyper-parameters. The authors also develop a hybrid optimization method combining a genetic algorithm with local search that can be used to optimize CNN structure. 

To improve CNN performance, some studies have explored the possibility of training multiple CNNs at the same time using different means to promote cooperation and specialization among them. Such sets of CNNs are called committees. \citet{Bochinski2017} propose a way of defining the CNN structure in terms of its hyper-parameters and a framework to automatically find out the best set of hyper-parameters using an evolutionary optimization algorithm. Additionally, they extend their framework to optimize a CNN committee. The goal of this study is to establish a framework to optimize CNN committees for better performance.

\citet{YananBing2019} are inspired by the success of ResNet and DenseNet and propose a genetic algorithm-based evolutionary approach of automatically designing CNNs using blocks from ResNet and DenseNet.The proposed algorithm is self-sufficiently automatic and does not expect any specific domain expertise from the user. 

\citet{XieYuille2017} venture to find a way to  automatically build effective CNN structures for a given dataset. As the number of possible network structures increase exponentially with the number of layers in the network, the authors employ a genetic algorithm to navigate through the expanding search space. They propose a fixed length encoding strategy which represents a network architecture in the population and a genetic algorithm that operates on this population to produce better generations. Experiments with the CIFAR10 and ILSVRC2012 datasets show that their method produces CNNs with competitive or better recognition accuracy. 

\citet{LeeKim2021} propose a genetic algorithm that considers CNN structure and its hyper-parameters both in the optimization space and produces a CNN with optimal architecture and hyper-parameter values for the given dataset. The performance of the algorithm is evaluated with  18F-Florbetaben Amyloid PET/CT images for classification of Alzheimer's disease. 

\citet{Loussaief2018} observe that hyper-parameters of a network such as network depth, number of filters, and their sizes dramatically affect the performance of a CNN. They propose a genetic algorithm that  finds the optimal values of those parameters for a given dataset, and thus produces an optimal CNN architecture. 

\citet{TianChen2018} propose a new genetic algorithm to find out the best suited pre-trained model for different datasets. They come up with a new genetic encoding model that represents different pre-trained CNN models in the population and an evolutionary approach to promote the best performing models in each generation. Experimental results have evidently shown that their approach outperforms some of the existing classification methods. 

\citet{CaiLuo2021} propose neural architecture search which is a resource-heavy search mechanism that automatically searches for CNN architectures. 
The authors devise an evolutionary framework that employs a multi-task, multi-objective search approach to find optimally balanced CNN architecture for a given task. 

Finally, as discussed earlier, \citet{Nagae2022} propose a method to automatically select effective layers using a genetic algorithm for InceptionV3 network. The authors also coin a term called Optimal Transport Dataset Distance (OTDD) to quantitatively evaluate a particular layer's efficacy as an update layer. They use OTDD to estimate a layer's importance as an update layer and compare a layer's contribution to accuracy with its OTDD score to reveal that layer OTDD score is indicative of a layer's capability of detecting features from the target dataset. 

From the above discussion we see that although there are several works that utilize genetic algorithm to select the best setting of hyper-parameters and network architectures, to the best of our knowledge, there is no existing work that employs a genetic algorithm to select the best blocks to be updated in a transfer learning setting. Our  investigation presented in this paper has filled this gap in the literature.

\section{Conclusion}
\label{sec:conclusion}
In order to help general users of transfer learning select effective update layers in a deep CNN, we have proposed a genetic algorithm-based solution that automatically selects CNN blocks (i.e., groups of layers) for effective transfer learning. The proposed block selection algorithm select blocks instead of layers that results in less computational time requirement and yet yields similar or slightly better accuracy over the baseline method. 
Currently the proposed  block selection algorithm is only implemented for a popular CNN called EfficientNet. This algorithm may easily be extended to other types of CNNs. In addition, other metaheuristic algorithms can be investigated to better select the update layers and blocks. 


\bibliographystyle{cas-model2-names}

\bibliography{cas-refs}


\end{document}